\documentclass[letterpaper, conference]{IEEEtran}

\IEEEoverridecommandlockouts

\usepackage{cite}
\usepackage{amsmath,amssymb,amsfonts}
\usepackage{algorithmic}
\usepackage{graphicx}
\usepackage{subcaption}
\usepackage{textcomp}
\usepackage{xcolor}
\usepackage{comment}
\usepackage{enumitem}
\usepackage{soul}

\begin{document}

\title{Property-Aware Robot Object Manipulation:\\a Generative Approach}

\author{
\IEEEauthorblockN{Luca Garello\textbf{*}\IEEEauthorrefmark{2}\IEEEauthorrefmark{6}, Linda Lastrico\textbf{*}\IEEEauthorrefmark{2}\IEEEauthorrefmark{4}, Francesco Rea\IEEEauthorrefmark{4}, Fulvio Mastrogiovanni\IEEEauthorrefmark{2},\\ Nicoletta Noceti\IEEEauthorrefmark{6}, Alessandra Sciutti\IEEEauthorrefmark{3}}

\IEEEauthorblockA{\IEEEauthorrefmark{3}Cognitive Architecture for Collaborative Technologies Unit (CONTACT),\\ Italian Institute of Technology, Genoa, Italy}
\IEEEauthorblockA{\IEEEauthorrefmark{4}Robotics, Brain and Cognitive Science Department (RBCS),\\ Italian Institute of Technology, Genoa, Italy}
\IEEEauthorblockA{\IEEEauthorrefmark{2}
Department of Informatics, Bioengineering, Robotics, and Systems Engineering (DIBRIS),\\ University of Genoa, Italy}

\IEEEauthorblockA{\IEEEauthorrefmark{6}
MaLGa Center - DIBRIS, University of Genoa, Italy\\}

\textbf{*} Indicates equal contribution, email: luca.garello@iit.it, linda.lastrico@iit.it }

\maketitle

\begin{abstract}
When transporting an object, we unconsciously adapt our movement to its properties, for instance by slowing down when the item is fragile. The most relevant features of an object are immediately revealed to a human observer by the way the handling occurs, without any need for verbal description. It would greatly facilitate collaboration to enable humanoid robots to perform movements that convey similar intuitive cues to the observers. In this work, we focus on how to generate robot motion adapted to the hidden properties of the manipulated objects, such as their weight and fragility. We explore the possibility of leveraging Generative Adversarial Networks to synthesize new actions coherent with the properties of the object. The use of a generative approach allows us to create new and consistent motion patterns, without the need of collecting a large number of recorded human-led demonstrations. Besides, the informative content of the actions is preserved. Our results show that Generative  Adversarial Nets can be a powerful tool for the generation of novel and meaningful transportation actions, which result effectively modulated as a function of the object weight and the carefulness required in its handling.
\end{abstract}


\section{Introduction}
\label{sec:intro}

Humans greatly leverage non-verbal communication to exchange information. 
This can be intentional and explicit, as it happens for a pointing gesture, or implicit, when information is embedded for instance in the body posture, the gaze direction, or the attitudinal stance while performing a specific gesture \cite{Sandini2018}. 
In a social context, such kind of natural, implicit information can be purposely emphasized to make our actions or future intentions more readable by the partner \cite{signaling}. In human-robot collaboration settings, it would be relevant that also robots could master this form of communication, modulating their goal-oriented actions in order to make them more intuitively legible to the human partner \cite{legibility}.
In particular, if we consider a scenario where humans and robots interact with the same set of objects, a robot should be able to both infer object properties from the observation of how the partner manipulates them and to select and perform the suitable manipulation action.
In this paper we focus on the latter objective, aiming at enabling robots to generate movements whose kinematics are properly modulated as a function of object properties. This would help to achieve a two-fold outcome:
(i) each object would be handled accordingly to its features, therefore mimicking natural human behaviour, and 
(ii) the gesture performed by the robot would convey the same information \textit{about} the object features, being therefore more transparent and readable for the human partner.\\
Considering object manipulation, some of the features that influence the behavior are the object weight and its fragility, i.e., whether it requires to be handled with special care.\\
In natural lifting movements, the arm kinematics correlates with the object weight \cite{weightFlanagan, velWeight}. 
Humans, since childhood, can detect this regularity and intuitively infer the weight of an object, just by observing others lifting it \cite{sciutti:weightChildren}. Moreover, reproducing human lifting motion on a humanoid platform has been proved to be effective at conveying the same information \cite{sciutti:weight}. 
In the lifting and transporting of objects that require careful handling, humans consciously adapt their motor strategy to preserve the integrity of the object.
When addressing carefulness, recent results of our group showed that careful handling significantly influences human transportation actions, to the point that a classifier can effectively discriminate low vs. high carefulness levels from videos of the transports alone \cite{HFR}. The problem of detecting the object properties from human motion was addressed also by Duarte \textit{et al}. \cite{carefulnessBillard}, who focused on the \textit{Careful} (C) / \textit{Not Careful} (NC) manipulation of objects, and tried to model the two movement classes using Gaussian Mixture Models (GMMs) \cite{b2}. 
According to the mentioned studies, appreciable information about the properties of handled objects can be inferred from the velocity profile of the performed  movements. 
In a human-robot interaction scenario, if we want a humanoid robot to deal with a particular set of objects in a credible way for human observers, the velocity profiles of its end-effectors should be accordingly adjusted, as to be comparable to the ones a human would adopt while transporting the same items.\\
In this work, we propose the use of Generative Adversarial Networks (GANs) to  generate novel velocity profiles suitable for different object categories, after being trained with the corresponding features acquired during human object manipulations.\\
GANs have been widely applied to different fields of research and their popularity is mainly due to their ability to generate new realistic data after being trained on a set of real examples. 
Although GANs are widely applied to different applications mostly in the computer vision domain, their application in the generation of multivariate time series has been explored only to a limited extent \cite{GAN,mogren2016c,b3}. In particular, in the field of human-robot-interaction there are few examples of GANs applied to motion generation \cite{ishiguro,yang2019appgan}.
Human motion kinematics are an example of a time-series, from which it is possible to extract descriptive features like velocity, acceleration profile, or the curvature radius associated with the trajectory. 

We posit that generative models can be used to ``learn" such time-dependent patterns, where the model is not only tasked with capturing the distributions of features within each time point, but it should also capture the potentially complex dynamics of those variables across time \cite{GAN}.
The long-term goal of the proposed approach is to control the end-effector of the humanoid robot iCub according to the synthetic velocity profiles, without having to exactly replicate the human kinematics from examples. 
In perspective, the complete framework will exploit on-going work to enable a robot to infer relevant object characteristics simply by observing how human partners manipulate them, therefore closing an object-related perception-action loop. 

\begin{figure}[t!]
    \centering
    \includegraphics[width=0.475\textwidth]{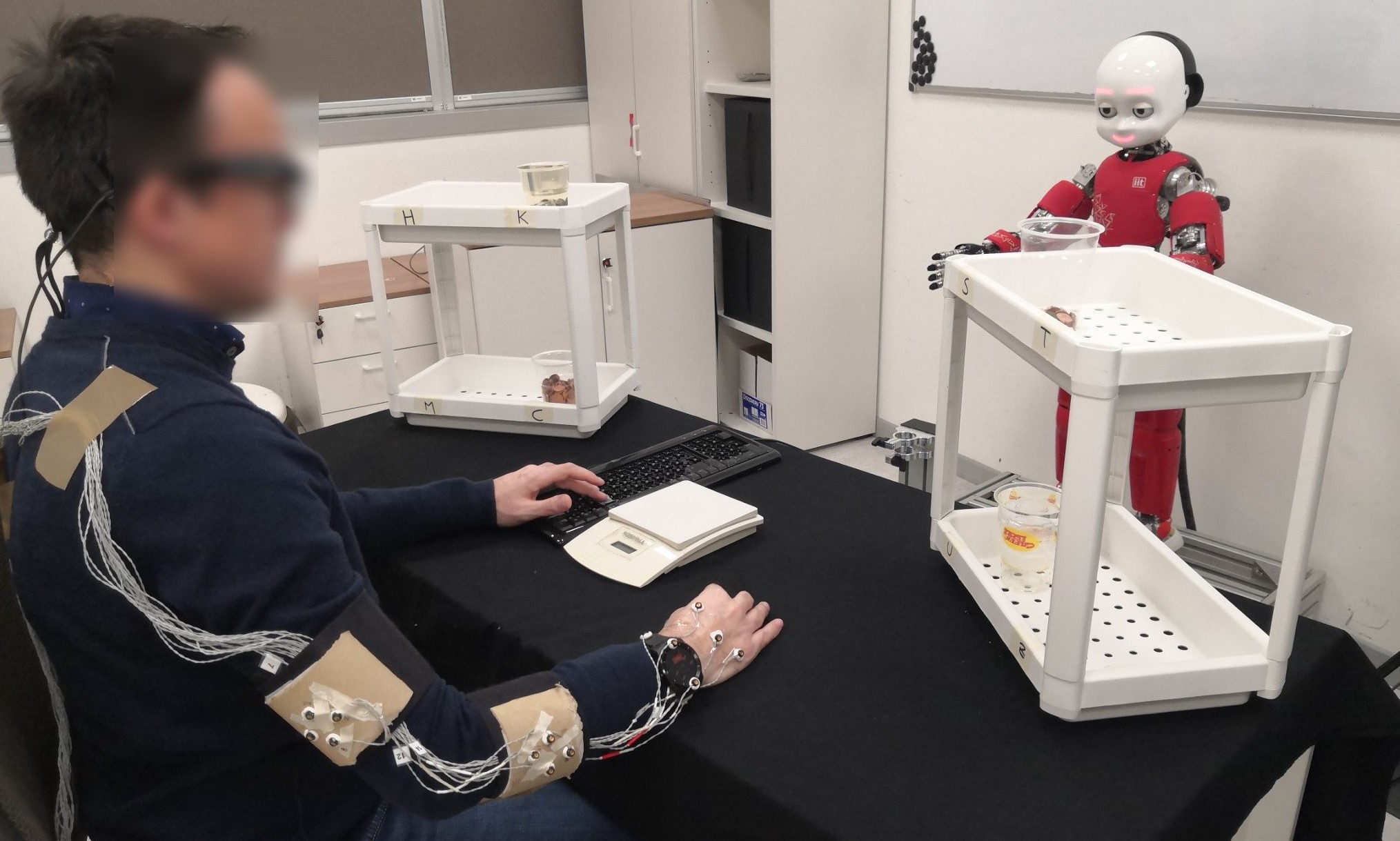}
    \caption{View of the experimental setup with a volunteer. The two shelves with the four glasses on them and the motion capture markers on the right hand are visible.}
    \label{fig:setup}
\end{figure}

\section{Collected Dataset}
\label{sec:dataset}

In this study, in order to train the generative models we used data recorded for a previous experiment, whose detailed methods and results are available in \cite{HFR}. 
Sixteen volunteers, none of them directly involved in our research, performed a series of transportation movements of four transparent glasses. 
The kinematic of the movements was recorded with multiple sensors, but only data originating from a motion capture system have been used in this study.
Active infrared markers of an Optotrak Certus\textsuperscript{\textregistered}, NDI, motion capture system were placed on each subject's right hand. 
The glasses were identical in shape and appearance, however they differed for the weight (W1: 167 gr; W2: 667 gr) and content: two of them were filled with water till the brim, and therefore required particular care from the subject to be moved. 
In this way, it was possible to define four classes of actions, depending on the properties of the manipulated object, namely (W1-NC) light and not careful, (W1-C) light and careful, (W2-NC) heavy and not careful, (W2-C) heavy and careful. 
The setup is shown in Figure \ref{fig:setup}, and the experiment consisted in transporting the different glasses from the eight possible positions on the shelves (two for each level: upper-bottom, left-right) to the scale on the table or in the opposite direction.
The experimental setup intrinsically allowed for a great variability in the performed gestures: the trajectories could be directed towards the left or the right side, towards lower or higher shelves and the direction of the motion could be either abductive or adductive.
The dataset used in this work for training the GANs consisted of 1001 total transport movements, (W1-NC: 248, W2-NC: 251, W1-C: 254, W2-C: 248 trials), which were segmented according to a threshold equal to the 5\% of the velocity peaks.
The sampling rate of the acquisition was 22 Hz.

\section{Methods}
\label{sec:methods}

As mentioned above, the goal of this work is to design a method to automatically generate synthetic velocity profiles consistent with the ones generated by a person manipulating an object with specific properties. 
In order to obtain a first corroboration for our hypotheses, for each transport movement, we considered the norm of the velocity, computed as in \eqref{eq:Vnorm}, from the three velocity components $V_{x}(t),\,V_{y}(t),\,V_{z}(t)$ at each time instant. 
\begin{equation}
\label{eq:Vnorm}
V(t)=\sqrt{V_{x}(t)^{2}+V_{y}(t)^{2}+V_{z}(t)^{2}}.\\
\end{equation}
Considered the norm as a feature, we trained a TimeGAN\cite{GAN} on all the transportation actions in the dataset.
In particular, we trained four different models, one for each object class, generating as many velocity profiles as the ones fed to each GAN.
Since the duration of the movements was characterized by a limited variability, we decided to pad the time series in each class with zeros to the length of the longest sequence. 
Although more sophisticated approaches are possible, this simple solution seems adequate in our case.
This resulted in velocity profiles of length $55$ for (W1-NC), $50$ for (W2-NC), $127$ for (W1-C) and $131$ for (W2-C).
All the TimeGAN models were implemented with 3-layer Gate Recurrent Units (GRU), each one with $28$ neurons (an extensive explanation of the architecture is provided by \cite{GAN}). 
The training phase lasted $2000$ epochs, and was performed with batches of size $15$.

In order to assess the quality of the generated data, we computed different metrics, inspired by those proposed by \cite{GAN,b3}:

\begin{enumerate}[label={\Alph*.}]
    \item \textit{Data distribution:} the synthetic data population should be distributed to match real data samples.
    \item \textit{Data discriminating power:} 
    a classifier trained to discriminate among the different object typologies on synthetic data should perform equally well when tested on real data and vice versa (i.e., train-on-synthetic, test-on-real approach). 
    \item \textit{Features preservation:} from synthetic data it should be possible to extract kinematics features similar to the ones originating from real data.
\end{enumerate}

\subsection{Data Distribution}

In order to understand how the time series of the different classes could be globally visualized, we performed a manifold analysis to reduce their dimensions. 
This approach shows the inherent structure of the data by representing them in a two dimensional space, allowing us to qualitatively evaluate whether the generated data are consistent with the original examples. More specifically, flattening the temporal dimension, we computed Principal Component Analysis (PCA), that is a linear projection of the data along the directions of their maximum variance, and t-distributed Stochastic Neighbor Embedding (t-SNE), which  highlights non-linear local structures. 

\subsection{Data Discriminating Power}

In order to quantitatively measure the similarity between original and generated data for each category, we decided to assess the performance of a classifier in discriminating the velocity profiles associated with the transport movements of \textit{careful/not careful} and \textit{light/heavy} glasses.
To this aim, and referring to \cite{b3}, we introduce two evaluation methods. 
The first metric is called \textit{Train on Real, Test on Synthetic} (TRTS), meaning that the model is trained and validated on real data and then tested on artificial data to discriminate either the carefulness or the weight of the transported glasses. 
Mirror-wise, the second evaluation metric consists in training and validating the model on the synthetic dataset, then testing it on the original dataset, and it is referred to as \textit{Train on Synthetic, Test on Real} (TSTR).
For the sake of comparison, all the classification tasks were addressed by using the same Long Short-Term Memory (LSTM) model
with an architecture having an input layer, a bidirectional LSTM with 64 hidden units and 2 fully connected layers, followed by an output layer with softmax as activation function. 
All the sequences were restored to their original length, by removing the values under a velocity threshold of 0.005 m/s.
For each training session, we splitted the training data into Training Set \textit{plus} Validation Set by using a 5-folds cross-validation. 
In Table \ref{table:classAll} and Table \ref{table:classSubsets} we report in parentheses the performance on the Validation Set. 
The Test Set was instead considered as \textit{all real data} (TSTR) or \textit{all synthetic data} (TRTS). 

\subsection{Features Preservation}

Another interesting aspect that must be considered when generating new velocity examples is the features preservation. Even though deep learning is widely applied for its ability of extracting features in a unsupervised way, our goal is to generate data that preserve specific characteristics (features) of the movements. 
Indeed, as discussed in Section \ref{sec:intro}, implicit information about the object we are manipulating is naturally embedded in the kinematics of our movements. 
The final goal of generating synthetic velocity profiles inspired by human ones is to endow iCub with the same communicative ability while performing its movements.
Therefore, it seems paramount that synthetic data preserve those features associated with specific object properties. 

Similarly to \cite{carefulnessBillard}, we want to generate data preserving such characteristics as  \textit{Movement Duration} (MD),  \textit{Peak Amplitude} (PA), and  \textit{asymmetry} in the velocity profiles. 
This last parameter, which we will refer to as AD/MD, represents the Acceleration Duration (AD) in relation to the full movement duration, and is commonly used to describe arm movement kinematics \cite{asymmetry,asymmetry2}.
It gives a percentage value for how far the velocity peak is ahead of or behind in relation to the MD, therefore indicating a longer or shorter deceleration period.

\begin{figure}[h!]
    \centering
    \includegraphics[width=0.9\linewidth]{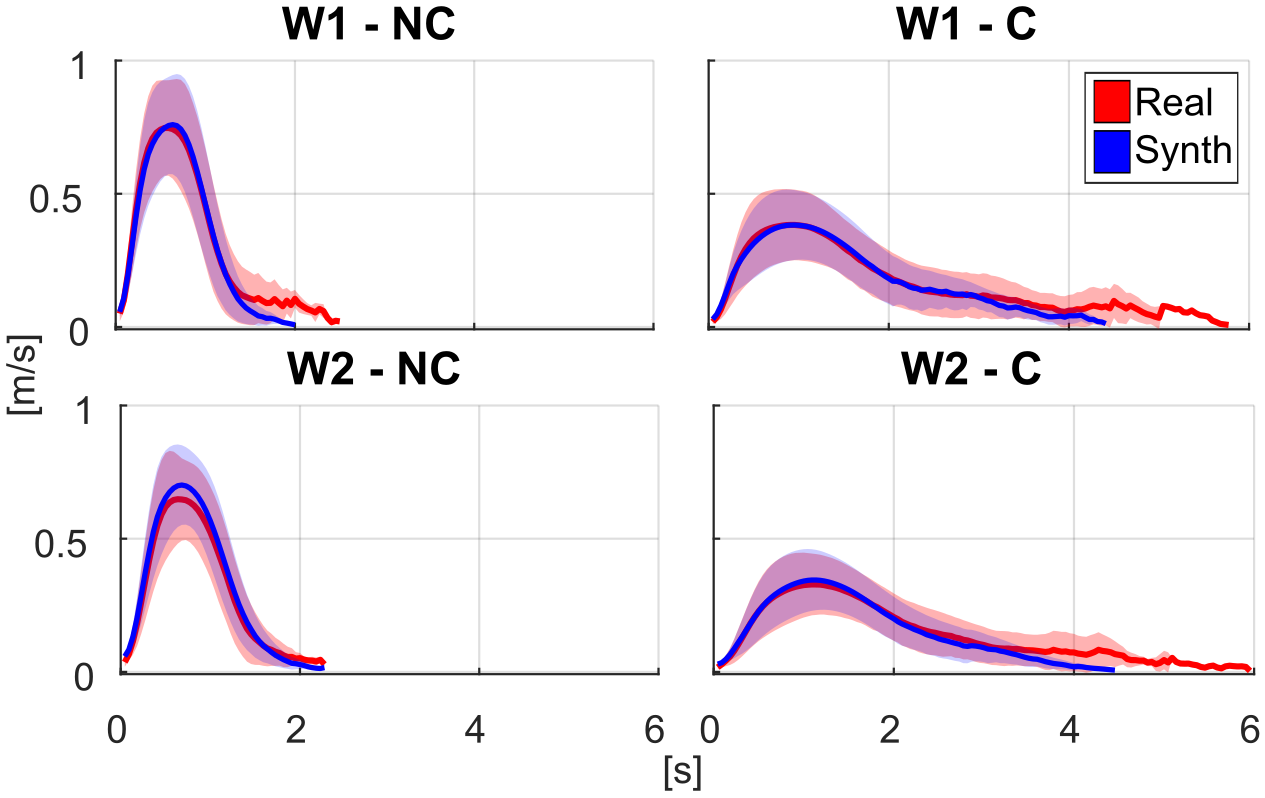}
    \caption{{Velocity profiles} of real (red) and synthetic (blue) data for each object class. The thick line represents the mean value, while the coloured area the standard deviation.}
    \label{fig:vel}
\end{figure}

\section{Results}
In Figure \ref{fig:vel} are shown global velocity profiles , which are obtained by calculating the mean and the standard deviation over the samples of every trial, for each of the four object classes. We have based all the subsequent analyses on these data. The tails visible on the original velocity profiles are due to an overall number of 21 trials (equal to $2.10\%$ of the entire dataset) which differed in the movement duration from the mean of each class (W1-NC: $1.44\pm0.17$, W2-NC: $1.61\pm0.19$, W1-C: $2.62\pm0.63$, W2-C: $3.04\pm0.69$ [s]) for more than three times the respective standard deviation. 
Therefore, such movements can be considered as outliers, and it is fair to assume that the GANs did not capture them, which is the reason why the generated velocity profiles (in blue) are more consistent with each other. 
\begin{figure*}[!ht]
    \centering
    \includegraphics[width=1\textwidth]{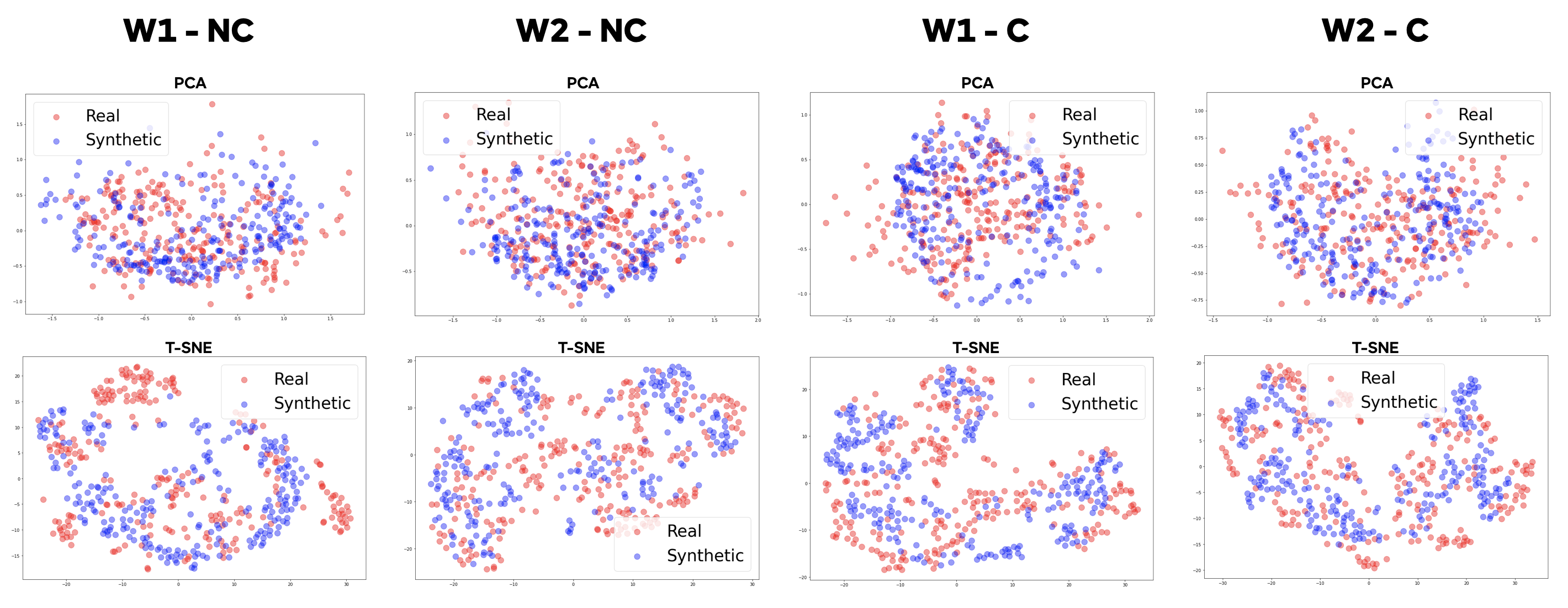}
    \caption{\textbf{Data distribution:} PCA and t-SNE projections for each movement category. The real and the synthetic data are superimposed.}
    \label{fig:pca/t-sne}
\end{figure*}
\begin{figure}[!h]
    \centering
    \includegraphics[width=0.475\textwidth]{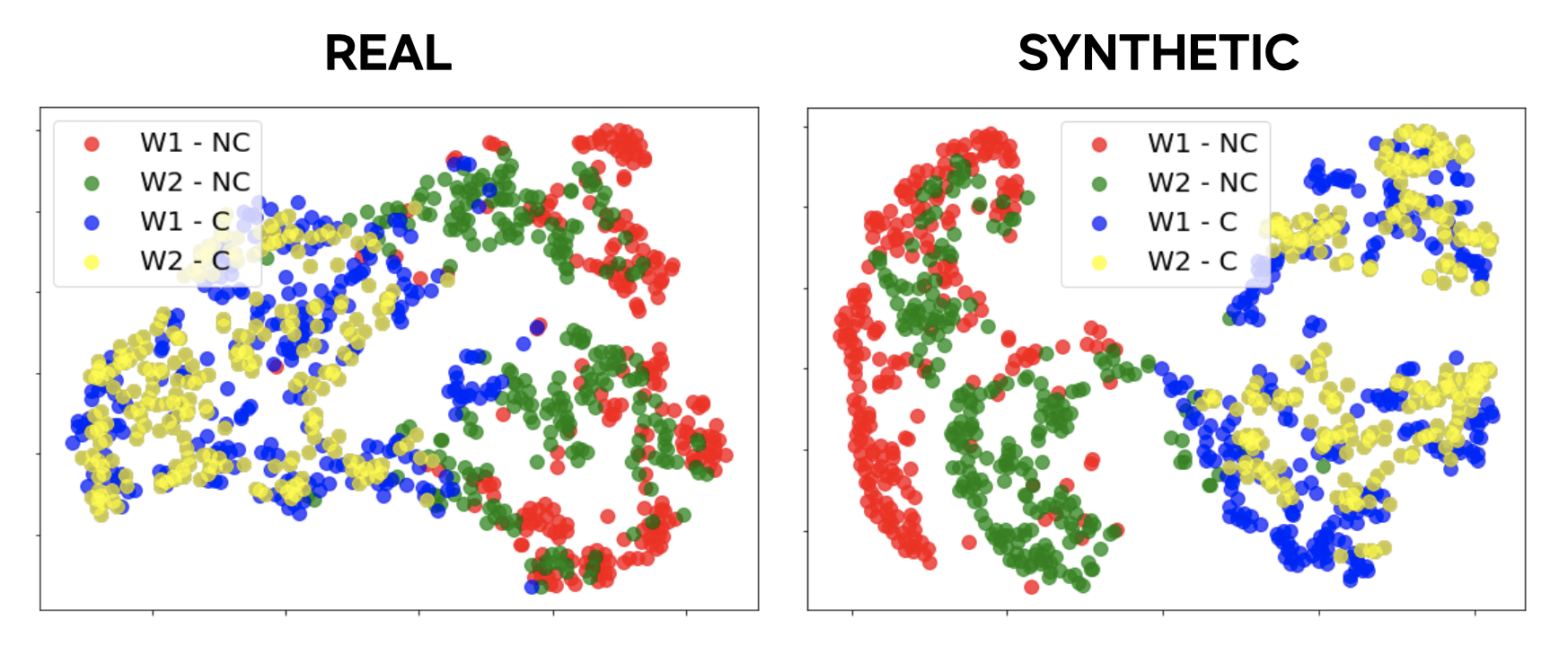}
    \caption{\textbf{t-SNE} representation for the two data sources. The four colors depict the glasses properties: red and green for the objects which did not require care (red - light, green - heavy), blue and yellow for those to be moved carefully (blue - light; yellow - heavy).}
    \label{fig:manifold}
\end{figure}
\subsection{Data Distribution}
In Figure \ref{fig:pca/t-sne} are shown Principal Component Analysis (PCA) and t-distributed Stochastic Neighbor Embedding (t-SNE) projections for real and synthetic data, divided according to the weight and the carefulness associated to the transported glasses. The synthetic data for each class are those generated by the corresponding GAN. This representation allows to appreciate how similar and overlapping are the original and the synthetic distributions.\\
In order to assess the separability among the different movements, we used again the t-SNE projection; the whole dataset (synthetic and real) is represented in Figure \ref{fig:manifold}, where the different colors highlight the four distinct classes of movements. This allows to evaluate which classes are more similar to each other according to their velocity profiles and whether the generated data reflect the same tendency. It can be noticed how the clusters linked to the care in the gestures are in general clearly separated, while those which differ only for the weight are largely overlapped, especially considering the portion of careful motions. Synthetic data appear even more strongly separable according to carefulness, and considering only the NC movements, the light (in red) and heavy (in green) objects seem to be distinguishable. 

\subsection{Data discriminating power}
The results reported in Tables \ref{table:classAll}, \ref{table:classSubsets} refer to multiple classification tasks performed using either the real data as training set (TRTS) or the synthetic data (TSTR) for the same purpose. As explained before, two main features were simultaneously characterising the handled glasses: the weight and the carefulness required by the presence of water. Even though the care in the manipulation seemed the prevalent feature, we decided to test the possibility of discriminating separately the weight or the care in the gestures. \\
We trained the same binary classifier model using the whole dataset (real or synthetic) to discriminate the carefulness or the weight, then testing it on the dual source of data. The corresponding accuracies are reported in Table \ref{table:classAll}. The performance are better when discriminating the carefulness (above $92\%$), with comparable results on both the test sets. The results on the weight classification are worse, however satisfying when training on real and testing on synthetic data, with an overall accuracy above $70\%$. The TSTR approach on the weight discrimination reveals a remarkable disparity between the accuracy obtained on the synthetic validation set ($97.40\pm1.08\%$) and the one on the real data used as test set ($62.76\pm1.01\%$).\\
{\renewcommand{\arraystretch}{1.4}
\begin{table}[!h]
\begin{center}
    \begin{tabular}{|c|c|c|}
    \hline
    & \shortstack{All \\ \textbf{Heavy/Light}} & \shortstack{All \\\textbf{Careful/Not Careful}}  \\ 
    \hline
    \hline
    TSTR & \shortstack{($97.40\pm1.08\%$)\\$62.76\pm1.01\%$} & \shortstack{($98.10\pm1.64\%$)\\$92.97\pm1.36$\%}  \\ 
    \hline
    TRTS & \shortstack{($70.53\pm3.04\%$)\\$75.18\pm6.17\%$} & \shortstack{($96.40\pm1.92\%$)\\$92.53\pm1.57\%$} \\ 
    \hline
    \end{tabular}
    \caption{Accuracy values training a classifier on real data and then testing it on synthetic data (TRTS) and vice versa (TSTR). The validation accuracy is reported in brackets, the test accuracy is below.}
    \label{table:classAll}
\end{center}
\end{table}
}
{\renewcommand{\arraystretch}{1.4}
\begin{table}[h]
\begin{center}
    \begin{tabular}{|c|c|c|}
    \hline
    & \shortstack{Heavy \\ \textbf{Careful/Not Careful}} & \shortstack{Light \\\textbf{Careful/Not Careful}}  \\ 
    \hline
    \hline
    TSTR & \shortstack{($99.0\pm0.71\%$)\\$87.29\pm4.42$\%} & \shortstack{($98.21\pm0.84\%$)\\$94.13\pm2.31$\%} \\
    \hline
    TRTS & \shortstack{$(96.99\pm2.36\%$)\\$94.15\pm3.17\%$} & \shortstack{($96.81\pm1.31\%$)\\$93.47\pm1.07\%$}  \\ 
    \hline
    \end{tabular}\vspace{4mm}
    
    \begin{tabular}{|c|c|c|}
    \hline
    & \shortstack{Careful \\ \textbf{Heavy/Light}} & \shortstack{Not Careful \\\textbf{  Heavy/Light  }}  \\ 
    \hline
    \hline
    TSTR & \shortstack{($87.65\pm3.91\%$)\\$61.39\pm5.29\%$} & \shortstack{($95.99\pm2.36\%$)\\$73.03\pm3.05\%$} \\
    \hline
    TRTS & \shortstack{($71.51\pm2.60\%$)\\$74.74\pm5.41\%$} & \shortstack{($78.35\pm3.89\%$)\\$79.60\pm6.77\%$} \\ 
    \hline
    \end{tabular}
\end{center}
\caption{TRTS and TSTR analysis over all the possible subclasses: in the upper part are the accuracies when discriminating the carefulness considering separately only the heavy or the light objects sub-samples. Likewise, in the lower part are the values when discriminating the weight either on the C or on the NC movements. The validation accuracy is in parentheses, the test accuracy is below.}
\label{table:classSubsets}
\end{table}
}
We also assessed the separability using as training set a halved dataset, in order to reduce a variability factor: the ability to discriminate the weight was evaluated using separately only the careful or the non careful velocity profiles; symmetrically, the possibility of classifying the carefulness was verified exploiting only the heavy or the light glasses. Table \ref{table:classSubsets} shows the matching results.
In particular, concerning the carefulness classification, the mean accuracy slightly improved for every combination, except for the TSTR when using only the simulated manipulations of heavy glasses. When discriminating the weight relying on the subset of not careful movements the performances improved, in particular for the TRTS, reaching a $79.60\pm6.77\%$ accuracy; the most critical result, even though above chance level ($61.39\pm5.29\%$), is the one related to the weight discrimination on the careful subset when training on synthetic data.
\begin{figure}[h]
    \centering
    \begin{subfigure}[b]{.48\textwidth}
    \includegraphics[width=1\linewidth]{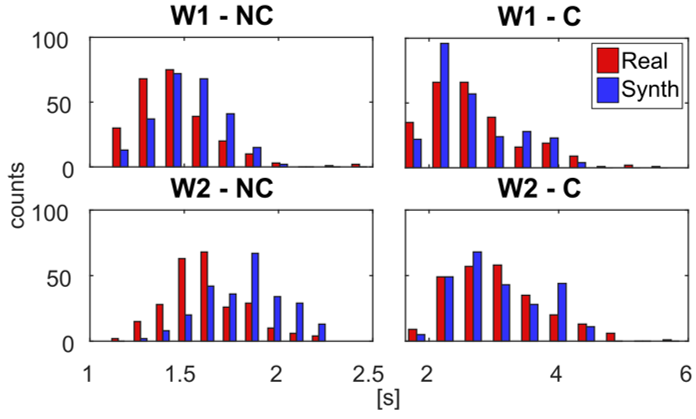}
    \caption{Movement Duration}
    \label{fig:MD}
    \end{subfigure}
\begin{subfigure}[b]{.48\textwidth}
    \centering
    \includegraphics[width=1\linewidth]{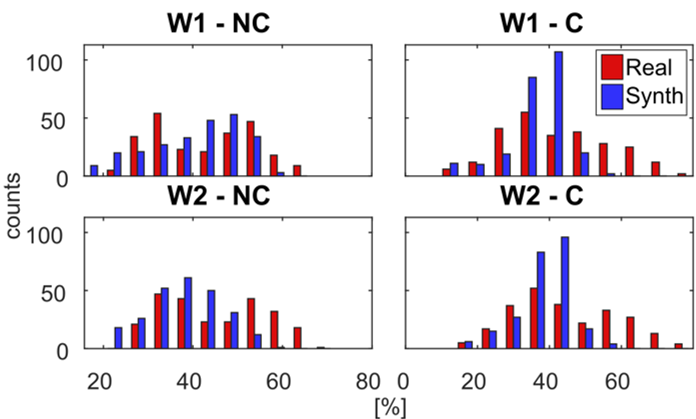}
    \caption{Acceleration Duration / Movement Duration}
    \label{fig:ADMD}
\end{subfigure}
\begin{subfigure}[b]{.48\textwidth}
    \centering
    \includegraphics[width=1\linewidth]{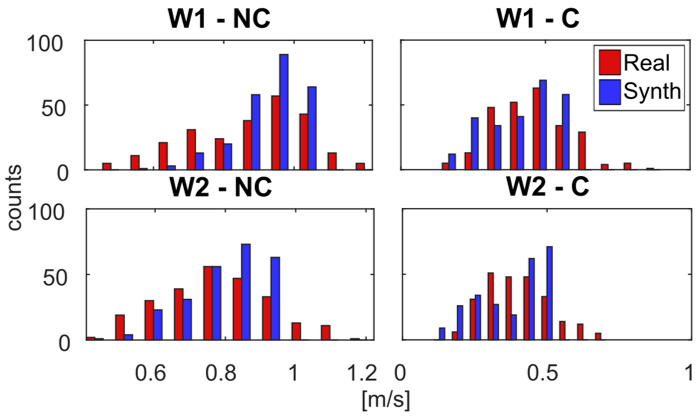}
    \caption{Velocity Peak Amplitude}
    \label{fig:vPeak}
\end{subfigure}
\caption{\textbf{Kinematics features distribution} for each object category for the Real (in red) and Synthetic (in blue) data.}
\label{fig:distributions}
\end{figure}
\subsection{Features preservation}
The last analysis we performed is meant to evaluate the persistence of meaningful properties. We assessed how close the real and synthetic distributions were according to some parameter characteristics of the kinematics of the movement: the MD, the AD/MD (i.e., the asymmetry in the velocity profile) and the Peak Amplitude. We report in Figure \ref{fig:distributions} the histograms corresponding to each one of the parameters.\\
In detail, Figure \ref{fig:MD} shows the duration of the transport movements. This feature cannot be generally considered as significant to detect the carefulness in a gesture or the weight of the item, since it is strongly dependent on the length of the trajectories: a glass full of water can be moved of a very short path, therefore the duration of that careful movement can be shorter than the one corresponding to the transportation of an empty glass for a more extended trajectory. However, in our dataset the paths covered with the four glasses were altogether the same, therefore we can consider the MD as characteristic of the object type. For the same distance covered, the durations for careful movements are markedly longer.\\
Concerning the AD/MD parameter reported in Figure \ref{fig:ADMD}, the difference among the four classes is less striking; it can be noticed that for the careful gestures the distribution of the synthetic data is less wide than the one of the original dataset, presenting a marked peak slightly before the 50\% of the profile. This means that the generated data, in this case, tend to show a more symmetrical acceleration and deceleration phase. 
Finally, observing the histograms for the Velocity Peak Amplitude (Figure \ref{fig:vPeak}), it can be noticed how such parameter shows lower values in the careful manipulations and that the synthetic data are able to capture such a tendency.

\section{Discussion}
From a qualitative point of view it is possible to notice that the synthetic velocity profiles are generated consistently with the real data distribution. This is supported by the mean velocity profiles in Figure \ref{fig:vel}, by the linear and non-linear manifold analysis in Figure \ref{fig:pca/t-sne} and by the kinematic parameters distributions of Figure \ref{fig:distributions}.
However, it must be stressed that the purpose of our work is not aiming at a perfect overlapping, that would mean a trivial copy of the data, but we are interested in a coherent representation of the distribution of the original data. The scope is to allow in the future the autonomous generation of appropriately communicative manipulation movements by robots, such as the humanoid iCub. 
When all the classes are visualized using a t-SNE representation, in Figure \ref{fig:manifold}, it is possible to notice that both real and synthetic data share the same structure: careful actions (\textbf{C}) and not careful actions (\textbf{NC}) appear to be separable with overlapping in their sub-classes representing the carrying of light (\textbf{W1}) and heavy (\textbf{W2}) objects.
It is worth considering that each class was generated independently from the others, training the GAN model using only the corresponding subset of original trials, without any knowledge on the velocity profiles of the other classes. The overlapping patterns emerging from the manifold analysis of the independently generated data strictly resemble the ones of the original dataset, meaning that the information was embedded in the data themselves. The proposed GANs were able to capture the characteristics of the original dataset, without any need of forcing the learning in a particular direction. \\
The difference between \textbf{C}/\textbf{NC} classes in synthetic data may be due to the fact that our model learned the most salient characteristics of these actions, overlooking the variability typical of the real data. A clearer separability seems to be present, according to the t-SNE representation of Figure \ref{fig:manifold}, even between the generated movements performed with heavy (green) or light (red) objects in the NC class. It is reasonable to assume that the original transportation movements were largely influenced by the presence or absence of water inside the glasses; when this ``disturbing'' factor was not present, that is when the gestures were not careful, it was the weight which played a role in influencing the movement kinematics. The GANs have captured this occurring, and the corresponding W1-NC and W2-NC velocity profiles are less overlapped in the synthetic manifold representation. The easier discrimination of weight when the gestures are not careful is also supported by the classifier results presented in Table \ref{table:classSubsets}. As suggested by the t-SNE representation, the easier separability between light and heavy objects in the synthetic dataset results in a validation accuracy of $95.99\pm2.36\%$, that drops to $73.03\pm3.05\%$ when testing the model on the NC original data, which are more overlapped. After these considerations, we hypothesize that the real actions embed a degree of variability that is not descriptive of the class, but which may be ascribed to the natural variety in how humans perform a repeated gesture. Indeed, Table \ref{table:classAll} and Table \ref{table:classSubsets} show in general how models trained on real samples are robust when tested on synthetic actions. When the classifier is instead trained on the synthetic data, there is a drop in the test accuracy over original data, that are naturally more noisy. This claim is supported also by the distributions of the kinematics features in Figure \ref{fig:distributions}, where it can be noticed how the real data present a broader base with less pronounced peaks.

\section{Conclusions}
We have presented how Generative Adversarial Networks can be a powerful tool for the generation of new and meaningful velocity profiles that characterize transportation actions of objects with different properties. This approach demonstrates that it is possible to generate actions that are consistent with the original data used to train the GANs and paves the way to future study to further investigate how such actions can be modulated.\\
Since the long-term goal of our work is to use the synthetic velocity profiles to control the motion of the iCub arm, it should be noticed that the proposed approach aims to enrich the robot's movements with a communicative intention. Indeed, even a child-shaped robot as iCub would be able to move without difficulties objects in the range of weights we considered. However, our purpose is to deliberately design legible movements to improve human-robot interaction. In this context, the use of GANs to reproduce in an original way the characteristics of human kinematics during different objects manipulations seems appropriate: it will allow to preserve also in the robot motion the implicit information on the properties of the handled object, making iCub movements not only effective but also readable and informative for the human partner. As a proof of concept, we recorded in simulation some examples of the humanoid robot moving its arm replicating velocity profiles corresponding to different classes of objects, comparing original and generated data. The video can be seen in the additional materials. This was done to verify, from a purely qualitative point of view, if the produced motions were plausible, comparable between synthetic and original data sources and to give an example of how the carefulness level and the weight influence the action, also on the robot. However, an extensive evaluation on how to exploit the synthetic profiles to generate meaningful movements is left as future work.\\ Finally, in this exploratory work we considered actions recorded with a motion capture setup, but in the future it will be useful to replicate this pipeline by using velocity profiles derived from the images acquired with the robot cameras. By doing so, it will be possible to achieve a more natural human-robot interaction.

\section*{ACKNOWLEDGMENT}
Alessandra Sciutti is supported by a Starting Grant from the European Research Council (ERC) under the European Union’s Horizon 2020 research and innovation programme. G.A. No 804388, wHiSPER
\bibliographystyle{IEEEtran} 
\bibliography{references}


\end{document}